\documentclass[letterpaper, 10 pt, conference]{ieeeconf}
\usepackage{times}
\usepackage[pdftex]{graphicx}
\usepackage{subfigure}
\usepackage{amsmath,amssymb,amsopn,amstext,amsfonts}
\usepackage{cancel}
\usepackage[space]{cite}
\usepackage{balance}
\usepackage{color}
\usepackage{mathtools}
\usepackage{multirow}
\usepackage{makecell}
\usepackage{algorithm}  
\usepackage{algorithmicx}  
\usepackage{algpseudocode} 
\usepackage{bm}

\usepackage{diagbox}
\usepackage{float}
\usepackage{epstopdf}
\usepackage{pifont}
\usepackage{multirow}
\usepackage{url}
\usepackage{tabularx}
\usepackage{booktabs}
\usepackage[linkcolor=black,citecolor=black,urlcolor=black,colorlinks=true]{hyperref}
\usepackage{verbatim}
\usepackage[skip=3pt, font={small}]{caption} 

\graphicspath{{../figure/}}
\DeclareGraphicsExtensions{.png,.jpg,.eps,.pdf}
\IEEEoverridecommandlockouts
\title{\LARGE \bf From Local Whole-Body VLA Behaviors to Scene-Scale\\ Aerial Manipulation}

\author{Weixiang Guo$^{*}$, Rui Jin$^{*}$, Haotian Jin, Xinhang Xu, Ruiyang Liu, Haoran Zhao,\\ Yi Wang, Weiqi Gai, Kun Cao, and Lihua Xie$^\dagger$
    \thanks{$^*$Equal contribution.}
    \thanks{$^\dagger$Corresponding author: {\tt\small elhxie@ntu.edu.sg}.}
    \thanks{
    Weixiang Guo, Rui Jin, Haotian Jin, Xinhang Xu, Ruiyang Liu, and Haoran Zhao are with the School of Electrical and Electronic Engineering, Nanyang Technological University, Singapore 639798.
    Yi Wang and Kun Cao are with the College of Electronics and Information Engineering, Shanghai Institute of Intelligent Science and Technology, Tongji University, Shanghai 201804, China.
    Weiqi Gai is with the School of Aeronautic Science and Engineering, Beihang University, Beijing 100191, China.
    Lihua Xie is with the NTU--VinUni Joint Research Laboratory for Embodied AI and Robotics, School of Electrical and Electronic Engineering, Nanyang Technological University, Singapore 639798, and VinUniversity, Hanoi, Vietnam.
    }
}

\makeatletter
\let\@oldmaketitle\@maketitle
\renewcommand{\@maketitle}{\@oldmaketitle
	\vspace{0.3cm}
	\centering
	\setcounter{figure}{0}
	\begin{minipage}{1.0\linewidth}
		\includegraphics[width=1.0\textwidth]{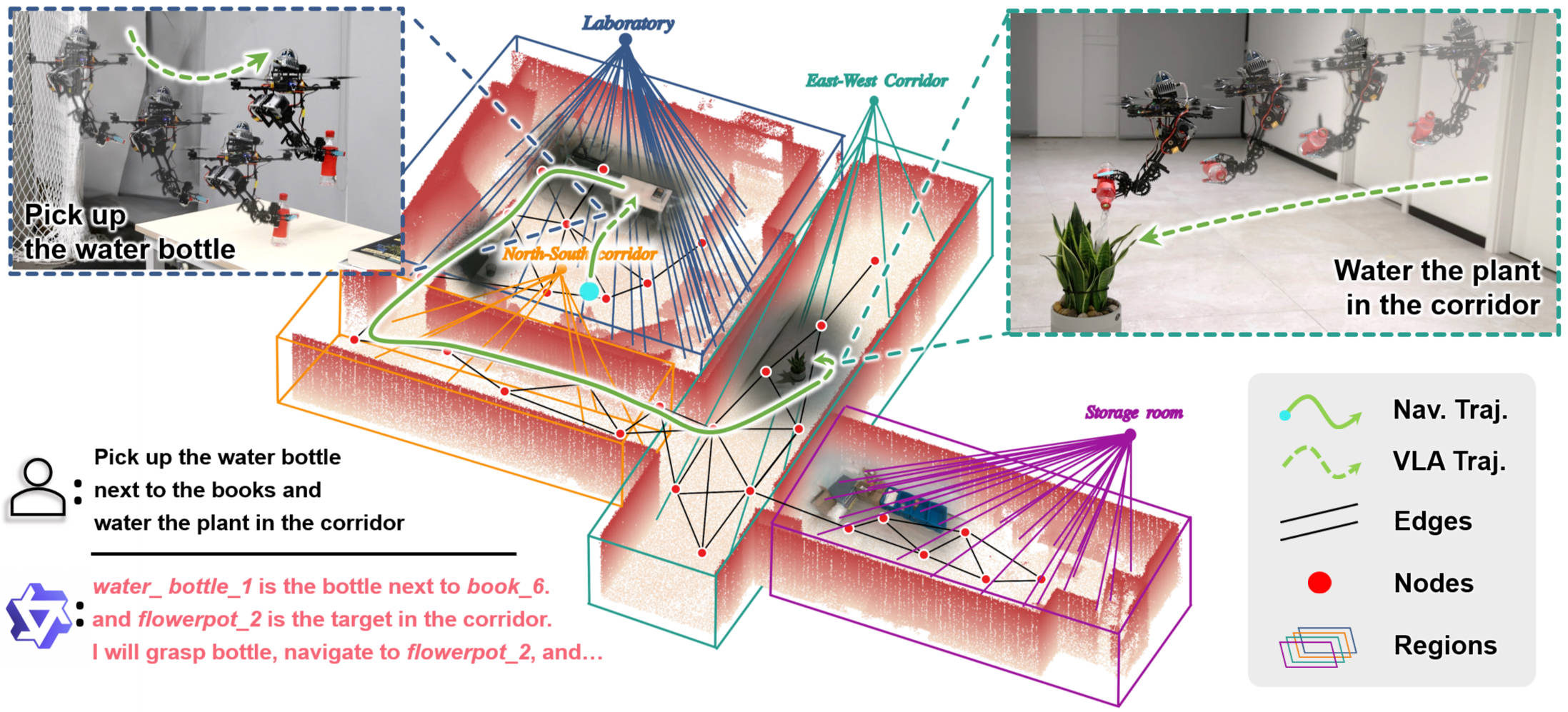}
		\vspace{-0.2cm}
        \captionof{figure}{\label{fig:teaser}The complete cross-site aerial manipulation process of picking up a water bottle and watering a plant. The language instruction is grounded to target object instances and feasible interaction regions through the relational Scene Graph. Solid and dashed green curves denote topology-guided navigation trajectories and local whole-body VLA trajectories, respectively. The left and right insets show the local VLA behaviors for bottle grasping and plant watering.}
	\end{minipage}
	\vspace{-0.3cm}
}

\makeatother

\begin{document}

\maketitle
\thispagestyle{empty}
\pagestyle{empty}

  \begin{abstract}

Vision-language-action (VLA) models enable task-conditioned interaction, but extending them to scene-scale aerial manipulation remains challenging due to costly whole-body demonstrations, latency-induced action–state misalignment, and cross-site behavior composition. We present a unified framework for synthetic policy training and scene-scale execution on articulated uncrewed aerial manipulators (UAMs). A scene-reconfigurable pipeline synthesizes task-conditioned, kinodynamically feasible trajectories and synchronized multiview observations for VLA training without physical-platform demonstrations. Measured-progress-aligned realization (MPAR) aligns asynchronously returned action chunks with measured execution progress and realizes them as continuous, dynamically feasible trajectories. A relational Scene Graph grounds language goals to object instances and feasible interaction regions, while topology-guided transfer connects local behaviors across sites. Local VLA skills achieve 39/60 successes (65.0\%) in simulation under oracle target and feasible-handoff conditions. Under 500-ms added latency, with and without a transient command-update stall, MPAR reduces median takeover phase error by 0.212~s over nominal-time alignment. The complete system completes 21/50 simulated multi-site missions (42.0\%) and is further validated on a physical articulated UAM.

\end{abstract}

\section{Introduction}
\label{sec:intro}
Vision-language-action (VLA) models map visual observations, language instructions, and proprioception to robot actions~\cite{zitkovich2023rt2,kim2024openvla}, drawing on sensorimotor experience across tasks and embodiments~\cite{openx2024}.
Recent work has begun extending this paradigm to aerial manipulation~\cite{sun2026airvla,tucker2026makeitfly}.
For an uncrewed aerial manipulator (UAM)~\cite{lin2025float,ductam,jinjie2026tro,tethered}, a VLA policy offers a way to coordinate the aerial base, arm, and gripper for language-specified interactions. Aerial mobility also makes it possible to perform these interactions at spatially separated sites without traversable ground connections. Realizing this capability requires extending local whole-body behaviors into a coordinated scene-scale mission (Fig.~\ref{fig:teaser}).

This extension presents three challenges. First, collecting diverse task-conditioned demonstrations on physical UAMs is costly and risky because base–arm–gripper motions must satisfy coupled dynamics and actuation limits. Second, inference and execution proceed asynchronously: when a new action chunk arrives, elapsed time may no longer indicate how far the UAM has actually progressed. The chunk must therefore be aligned with the physical execution state and incorporated without command discontinuities or violations of the platform's motion limits. Third, a relational language goal must be resolved not only to the intended object instance but also to a feasible whole-body handoff state from which the local behavior can begin. Reaching such states across spatially separated interaction sites additionally requires collision-free, kinodynamically feasible transfer.

To address these challenges, we propose a unified framework
that learns local whole-body VLA behaviors and composes them
into scene-scale aerial missions. Rather than assigning an
entire mission to a single policy, we retain learned local
coordination while explicitly handling physical realization
and cross-site transfer. We evaluate the framework in simulation
and through representative physical trials.
Our main contributions are:

\begin{itemize}
    \item A scene-reconfigurable supervision pipeline combining
    kinodynamically feasible whole-body expert planning with
    synchronized multiview rendering to generate task-conditioned
    training data for physical deployment without demonstrations
    collected on the target UAM.

    \item Measured-progress-aligned realization (MPAR), which
    projects estimated handoff states onto incoming VLA paths,
    anchors continuous takeover to active command states, and
    jointly optimizes path progress and trajectory timing under
    whole-body constraints.

    \item A scene-graph-based composition interface that resolves
    relational targets, derives feasible whole-body handoff states
    from skill-conditioned interaction regions, and guides
    cross-site transfer using polyhedral topology.
\end{itemize}

\section{Related Work}
\label{sec:related_works}
\subsection{Vision-Language-Action Models for Aerial Manipulation}
\label{subsec:aerial_vla_learning}
VLA research for aerial robots spans language-conditioned navigation~\cite{sun2026autofly}, simulated aerial manipulation~\cite{sun2026airvla}, and deployment on physical platforms~\cite{tucker2026makeitfly}. Existing VLA and visuomotor policies acquire supervision through simulator teleoperation, generated navigation trajectories, Gaussian-splatting-based augmentation, and handheld cross-embodiment demonstrations~\cite{sun2026airvla,sun2026autofly,tucker2026makeitfly,gupta2025umi}. These approaches either retain dependence on human demonstrations, focus primarily on navigation, augment visual appearance without generating new whole-body interactions, or express actions in an end-effector-centric space. For articulated UAMs, this leaves an important gap: automatically generating task-conditioned supervision that is both whole-body and kinodynamically feasible, without collecting demonstrations on the platform. 

Physical deployment introduces a distinct challenge: converting learned action predictions into reliable robot motion. AirVLA~\cite{tucker2026makeitfly} deploys a pretrained VLA on a physical quadrotor equipped with a gripper, using RTC~\cite{black2025rtc} for asynchronous execution and payload-aware guidance. Flying Hand and UMI-on-Air demonstrate learned visuomotor control on physical articulated UAMs, with the latter incorporating embodiment-aware controller guidance during policy sampling, although neither employs a language-conditioned foundation VLA~\cite{he2025flying,gupta2025umi}. More general runtime methods address observation--execution mismatch and execution-time context prediction at the policy or action-sequence level~\cite{wang2026remac,jiang2026futurertc}, while model-based aerial planners generate dynamically feasible trajectories from specified references or goals~\cite{zhepei22tro,deng25tro}. These two lines address execution latency and motion feasibility largely separately. For articulated UAMs, reliably realizing rolling whole-body VLA action chunks under variable inference latency requires both alignment with actual execution progress and continuous, constraint-aware trajectory realization.

\subsection{Scene Graphs for Scene-Scale Aerial Manipulation}
\label{subsec:scene_graph}

Open-vocabulary 3D scene representations organize object instances and their spatial relations for context-aware retrieval and large-scale navigation~\cite{gu2024conceptgraphs,werby2024hovsg}. Building on these representations, SayPlan and DovSG support long-horizon task planning and combined navigation--manipulation through structured scene reasoning~\cite{rana2023sayplan,yan2025dovsg}. These systems primarily resolve language goals to semantic instances, regions, or high-level skills. 
For articulated UAMs, however, we additionally require grounding to an interaction region that admits a feasible whole-body state for initiating the local skill.

Beyond semantic grounding, hierarchical scene and topological graphs provide global connectivity. USS-Nav~\cite{gai2026ussnav} couples polyhedral connectivity, semantic regions, and object instances for LLM-augmented UAV navigation, but primarily supports symbolic reasoning or point-level navigation. Safe-flight-corridor and whole-body planners, however, generate collision-free, dynamically feasible motion from specified geometric targets~\cite{zhepei22tro,deng25tro, gsplanner,sketchprior}, but do not resolve relational language goals. Scene-scale aerial manipulation therefore requires bridging relational grounding with executable interaction-region selection and whole-body-feasible transfer.

\begin{figure*}[t]
    \centering
    \includegraphics[width=1.0\textwidth]{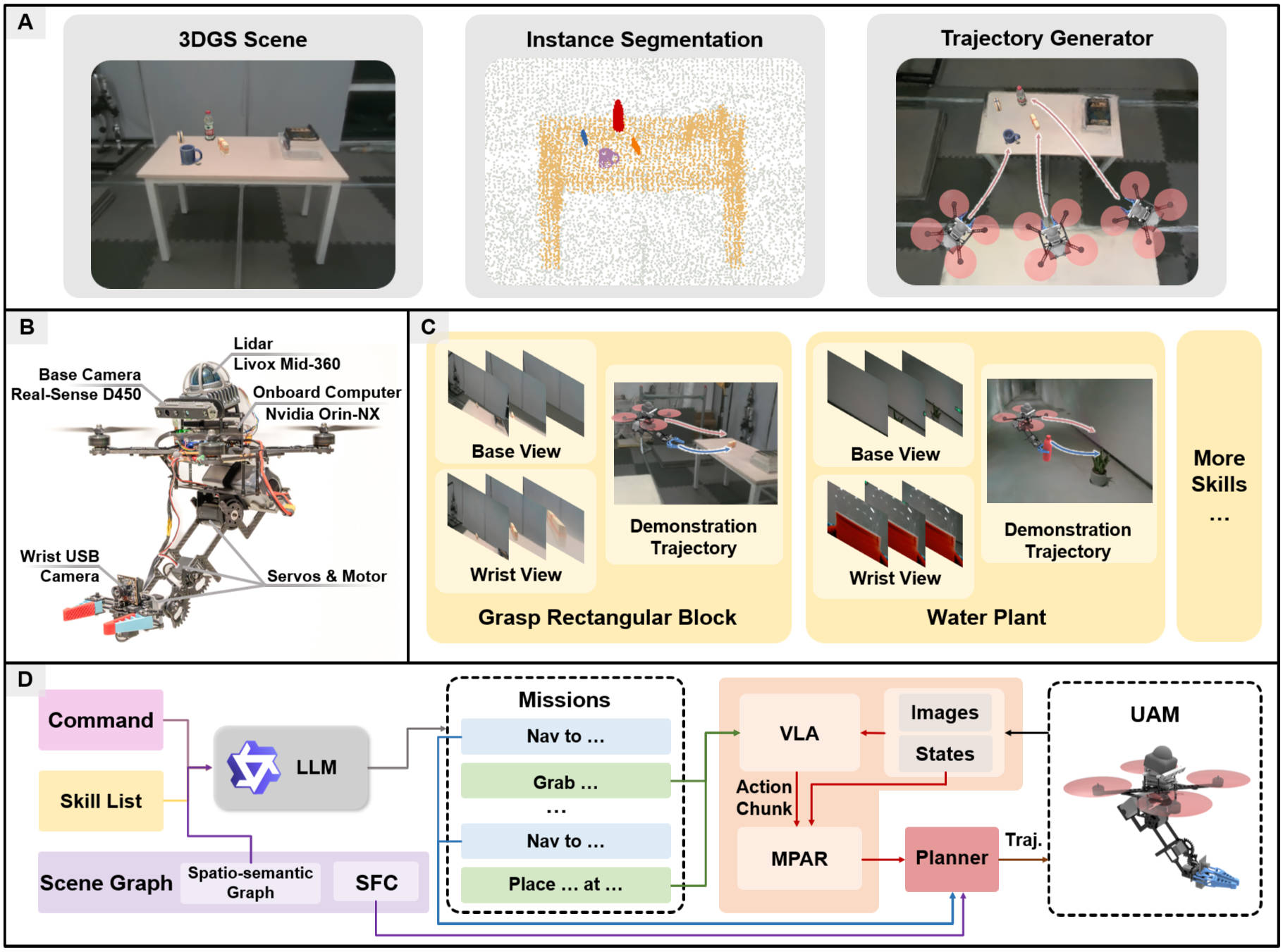}
    \caption{System overview. (A) Reconfigurable 3DGS-based supervision
    synthesis. (B) Physical articulated UAM. (C) Synthesized local-skill
    observations and expert trajectories. (D) Online execution combining
    Scene Graph grounding and transfer, whole-body VLA prediction, MPAR,
    and kinodynamic trajectory realization.}
    \label{fig:system_overview}
    \vspace{-0.3cm}
\end{figure*}

\section{System Overview}
\label{sec:system_overview}
Given a natural-language mission $\ell$ and a preconstructed Scene Graph $\mathcal{G}$ registered to a metric map, we consider an articulated UAM comprising an aerial base, a multi-joint arm, and a gripper.

Within this setting, as shown in Fig.~\ref{fig:system_overview}, the VLA is trained offline on synthesized whole-body demonstrations. At runtime, the mission layer queries $\mathcal{G}$ to resolve target instances and local instructions, while topology-guided transfer brings the UAM to a feasible handoff region. After a verified handoff, the VLA maps multiview images, proprioception, and the local instruction to finite-horizon base--arm--gripper action
chunks. MPAR uses the measured state to align asynchronously returned chunks and generate continuous, constraint-aware trajectories. Visual and proprioceptive feedback close the local execution loop, while skill outcomes advance the mission.

\section{Scene-Reconfigurable Whole-Body VLA Supervision}
\label{sec:supervision_synthesis}
To synthesize whole-body VLA supervision for physical-UAM deployment, we reconfigure each task at both the geometric and visual levels. For each episode, the pipeline samples the UAM initial state together with target and distractor identities and poses within task-valid ranges. It then derives an object-relative interaction goal from the sampled configuration and language instruction. A whole-body planner generates coordinated aerial-base, arm, and gripper references under motion, joint, continuity, and collision constraints, retaining feasible rollouts as expert behaviors.
Separate Gaussian representations of the static environment and movable assets are registered in a common metric frame, allowing assets to be independently placed with six-DoF poses while preserving the deployment-scene appearance. At each sampled time $t$, the pipeline renders synchronized base- and wrist-camera images from the expert state. These images, together with the whole-body state $\mathbf{x}_t$ and language instruction $\ell$, form the policy input, while the corresponding expert future action chunk $A_{t:t+H}$ serves as the supervision target.

We instantiate the policy by fine-tuning a pretrained $\pi_{0.5}$ model through OpenPI~\cite{piwebsite2025pi05}. Rank-16 and rank-32 LoRA adapters are applied to its PaliGemma language module and action expert, respectively. The model receives $224\!\times\!224$ base- and wrist-camera images, proprioception, and a language instruction, and predicts ten future actions at 0.25-s intervals (2.5-s horizon). At observation time $t$, the proprioceptive state and the $k$-th action are
\begin{equation*}
\begin{aligned}
\mathbf z_t &=[\boldsymbol{\theta}_t,\mathbf v_t^B,\phi_t,\vartheta_t]\in\mathbb R^8,\\
\mathbf a_{t,k} &=[\boldsymbol{\theta}_{t,k}^{\star},g_{t,k},\Delta\mathbf p_{t,k}^B,\Delta\psi_{t,k}]\in\mathbb R^8.
\end{aligned}
\end{equation*}
Here, $\boldsymbol{\theta}_t$ and $\boldsymbol{\theta}_{t,k}^{\star}$ are the current and target 3-DoF arm configurations, $\mathbf v_t^B$ is the base translational velocity in the yaw-aligned body frame, and $\phi_t$ and $\vartheta_t$ are roll and pitch. The action additionally contains a gripper-closing score $g_{t,k}$ and base-position and yaw offsets relative to the observation-time base pose. All waypoints in a chunk share this anchor rather than being accumulated recursively, and $g_{t,k}>0.5$ denotes closing. During fine-tuning, the joint targets are delta-encoded relative to $\boldsymbol{\theta}_t$ and converted back to absolute targets before runtime realization. Simulation and physical policies are separately fine-tuned on MuJoCo-collected and 3DGS-rendered data, respectively (39,200 windows each), with dataset-specific quantile normalization. AdamW is used in training for 30,000 updates with a batch size of 32 and cosine learning-rate decay from $2.5\!\times\!10^{-5}$ to $2.5\!\times\!10^{-6}$ after 1,000 warmup steps. Each policy uses its final step-29999 checkpoint, fixed a priori.

\section{Measured-Progress-Aligned Whole-Body Realization}
\label{sec:measured_progress_realization}
\begin{figure}[t]
    \centering
    \includegraphics[width=0.5\textwidth]{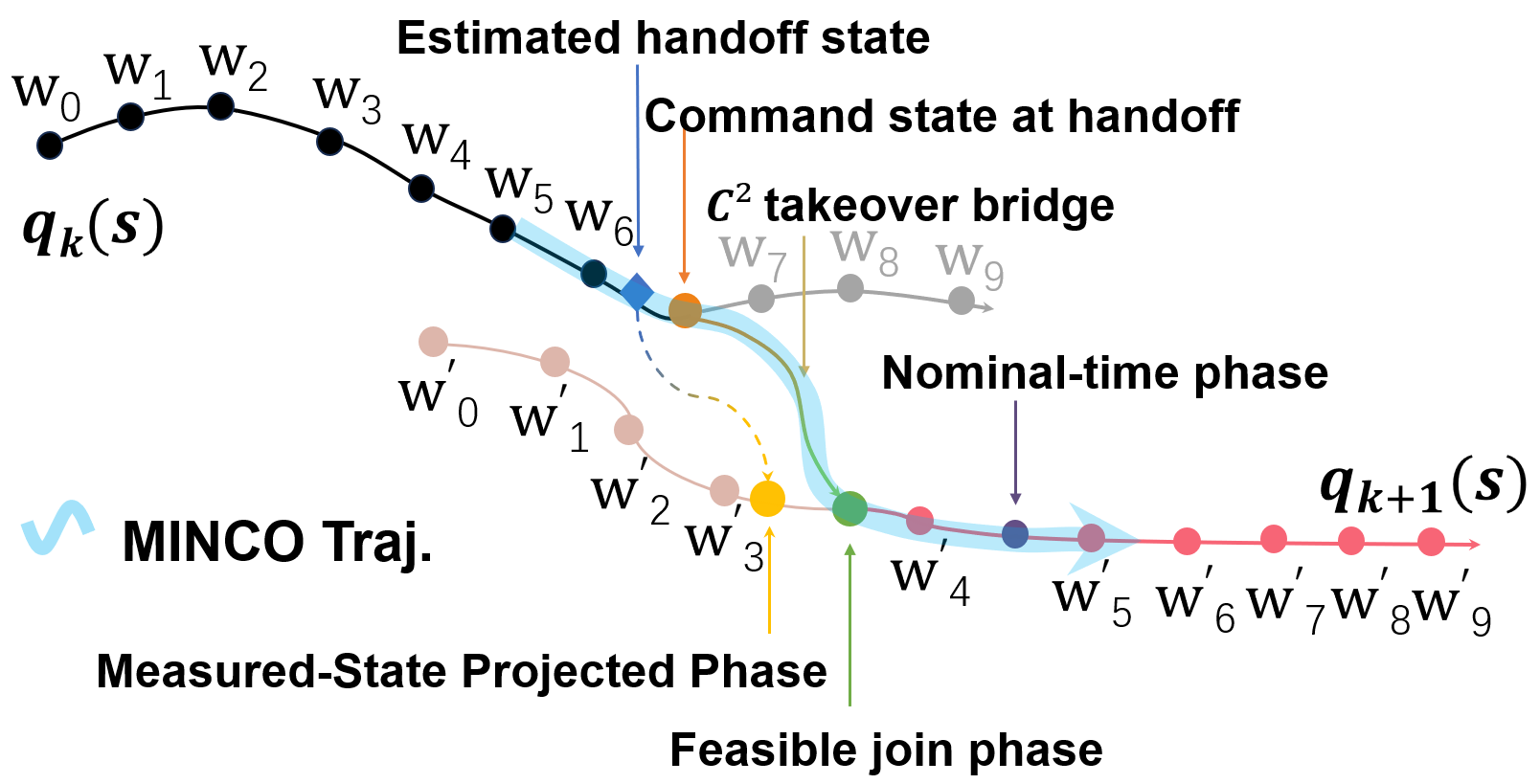}
    \caption{Measured-progress-aligned takeover between consecutive VLA chunks. Nominal-time alignment selects a phase according to elapsed time, whereas MPAR projects the estimated handoff state onto the incoming VLA path. Starting from the active command state at handoff, a $C^2$ bridge reaches a forward feasible join phase, after which MINCO realizes the remaining VLA-path suffix as a continuous trajectory.}
    \label{fig:progress_realization}
    \vspace{-0.5cm}
\end{figure}

\subsection{Whole-Body VLA Path Representation}
As illustrated in Fig.~\ref{fig:progress_realization}, at observation time $t_o$, the VLA predicts an action chunk $\mathcal{A}_o=\{\mathbf{a}_k\}_{k=0}^{K}$, which is decoded into synchronized whole-body waypoints $\mathbf{w}_k=(\mathbf{p}_k,\psi_k,\boldsymbol{\theta}_k,g_k,s_k)$. Here, $\mathbf{p}_k$, $\psi_k$, and $\boldsymbol{\theta}_k$ denote the aerial-base position, yaw, and arm configuration, respectively; $g_k$ is the gripper command; and $s_k=(k+1)\Delta t$ is the nominal offset from $t_o$, such that $k=0$ denotes the first future waypoint rather than a state at the observation time. Piecewise interpolation of the continuous components defines the geometric whole-body path
\begin{equation}
\mathbf{q}(s)
=
\begin{bmatrix}
\mathbf{p}^{\top}(s) &
\psi(s) &
\boldsymbol{\theta}^{\top}(s)
\end{bmatrix}^{\top},
\qquad s\in[s_0,s_K].
\label{eq:vla_whole_body_path}
\end{equation}
A shared phase $s$ indexes the base, yaw, and arm jointly, thereby preserving their coordination within the VLA-predicted behavior. Although $s_k$ inherits the nominal temporal spacing of the action chunk, $s$ is subsequently treated as path progress rather than a prescribed execution clock; its timing is determined during physical realization. Discrete gripper commands retain their source-waypoint associations and are mapped to the managed path after execution starts, triggering with retimed geometric progress rather than the original chunk timestamps.

\subsection{Measured-Progress Alignment and Continuous Takeover}
When a new action chunk is received at $t_r$, the runtime schedules its handoff at $t_h=\max\!\left(t_o+T_{\mathrm{target}},\,t_r+T_{\mathrm{lead}}\right),$
where $T_{\mathrm{lead}}$ reserves time for planning and validation. The currently committed trajectory remains active during inference, and $t_h$ is restricted to its valid horizon. Let $\mathbf{x}_m(t_r)$ and $\mathbf{x}_c^{-}(t_r)$ denote the measured and commanded whole-body states, respectively. Their residual is propagated over $\Delta t_h=t_h-t_r$ to estimate the state at handoff. For example,
\begin{equation}
\begin{aligned}
\hat{\mathbf p}_h
&=\mathbf p_c^{-}(t_h)+\Delta\mathbf p
 +\Delta\mathbf v\,\Delta t_h
 +\tfrac{1}{2}\Delta\mathbf a\,\Delta t_h^2,\\
\hat{\boldsymbol\theta}_h
&=\boldsymbol\theta_c^{-}(t_h)
 +\Delta\boldsymbol\theta
 +\Delta\dot{\boldsymbol\theta}\,\Delta t_h,
\end{aligned}
\label{eq:handoff_state_prediction}
\end{equation}
with analogous propagation for velocity and yaw.

The observation clock gives the nominal path phase $s_{\mathrm{nom}}=t_h-t_o$. We instead obtain a measured-aligned phase by projecting the estimated handoff state onto the fresh whole-body path within a bounded neighborhood $\mathcal{W}(s_{\mathrm{nom}})$:
\begin{equation}
\hat{s}
=
\underset{s\in\mathcal{W}(s_{\mathrm{nom}})}{\arg\min}\;
D\!\left(\hat{\mathbf x}_h,\mathbf q(s)\right),
\label{eq:measured_progress_projection}
\end{equation}
where \(D\) measures normalized disagreement between the predicted handoff state and the path configuration together with its phase-derived motion at \(s\). The correction is accepted only if it satisfies recovery bounds and improves agreement over $s_{\mathrm{nom}}$; otherwise, $\hat{s}\leftarrow s_{\mathrm{nom}}$.

For each valid handoff, let $\tilde{s}_h$ denote the phase selected online, where $\tilde{s}_h=s_{\mathrm{nom}}$ for nominal-time alignment and $\tilde{s}_h=\hat{s}$ for MPAR. We define
\begin{equation}
\begin{aligned}
s_h^\star
&=\underset{s\in[s_0,s_K]}{\arg\min}\;
  D\!\left(\mathbf{x}_m(t_h),\mathbf q(s)\right),\\
\delta s
&=\tilde{s}_h-s_h^\star,
\qquad e_s=|\delta s|,\\
L_p(a,b)
&=\int_{\min(a,b)}^{\max(a,b)}
  \|\mathbf p'(s)\|_2\,\mathrm ds,\\
\ell_{\mathrm{skip}}
&=\mathbb I[\delta s>0]L_p(s_h^\star,\tilde{s}_h),\\
\ell_{\mathrm{repeat}}
&=\mathbb I[\delta s<0]L_p(s_h^\star,\tilde{s}_h).
\end{aligned}
\label{eq:phase_metrics}
\end{equation}
Here, $s_h^\star$ is the post-hoc phase obtained by projecting the state measured at handoff onto the incoming VLA path. Since $s$ inherits the nominal waypoint timing, $e_s$ is measured in seconds; positive and negative $\delta s$ indicate skipped and repeated aerial-base path portions, respectively. Episode-level arc lengths are summed over valid handoffs.

From $\hat{s}$, the runtime searches for a feasible join phase $s_{\mathrm{join}}$ and bridge duration $T_b$. The bridge uses the exact state of the active command at $t_h$,
\begin{equation}
\mathbf b(0)=\mathbf x_c^{-}(t_h),
\qquad
\mathbf b(T_b)=\mathbf q(s_{\mathrm{join}}),
\label{eq:continuous_takeover}
\end{equation}
while matching the endpoint derivatives. Candidate transitions are rejected if they violate base velocity, acceleration, yaw-rate, or joint-rate limits. Importantly, $\hat{\mathbf x}_h$ is used only to identify the appropriate VLA-path suffix; retaining $\mathbf x_c^{-}(t_h)$ as the bridge boundary ensures a continuous command handoff. If no admissible takeover is found, the chunk is rejected while the committed trajectory remains active and replanning proceeds from a fresh observation.

\subsection{Progress-Parameterized MINCO Realization}
The bridge and remaining VLA-path suffix define the reference for constrained trajectory realization. Following the QuadHand~\cite{quadhand} whole-body planning framework, we parameterize the executable flat output
\begin{equation}
\mathbf z(t)
=
\begin{bmatrix}
\mathbf p^{\top}(t) &
\psi(t) &
\boldsymbol\theta^{\top}(t)
\end{bmatrix}^{\top}
\end{equation}
using $\mathfrak{T}_{\mathrm{MINCO}}^{4}$ for the aerial-base position and $\mathfrak{T}_{\mathrm{MINCO}}^{2}$ for yaw and each arm joint~\cite{zhepei22tro}. The initial boundary is fixed to the active command state at $t_h$, preserving position, velocity, and acceleration for the base and state and rate for yaw and the arm. Since the bridge serves only to establish this continuous takeover, VLA-path tracking begins after $s_{\mathrm{join}}$.

For the remaining $M$ trajectory intervals, we jointly optimize their durations $T_i$ and vertex phases $s_i$, using $\boldsymbol\eta\in\mathbb R^{M-1}$ and $\eta_{\rho}\in\mathbb R$ to parameterize monotonic progress:
\begin{equation}
\begin{aligned}
s_i &= s_{\mathrm{join}}
+\left(s_K-s_{\mathrm{join}}\right)
\rho\sum_{j=1}^{i}\pi_j, \\
\boldsymbol\pi
&=\operatorname{softmax}([\boldsymbol\eta^{\top},0]^{\top}),
\qquad \rho=\operatorname{sigmoid}(\eta_{\rho}).
\end{aligned}
\label{eq:monotonic_path_progress}
\end{equation}
which guarantees $s_{\mathrm{join}}<s_1<\cdots<s_M<s_K$. Together with the optimized durations, these phases determine the execution rate $\dot{s}\approx(s_i-s_{i-1})/T_i$, allowing the UAM to accelerate or slow its traversal without changing the ordering of the VLA behavior.

Denoting the MINCO intermediate waypoints by $\mathcal P$, the resulting optimization is written compactly as
\begin{equation}
\begin{aligned}
\min_{\mathcal P,\mathbf T,\boldsymbol\eta,\eta_{\rho}}
\quad
&J_{\mathrm{smooth}}
+J_{\mathrm{cont}}
+J_{\mathrm{prog}}
+J_{\mathrm{feas}},
\end{aligned}
\label{eq:progress_minco_objective}
\end{equation}
where $J_{\mathrm{smooth}}$ penalizes the high-order derivatives of the MINCO trajectories. $J_{\mathrm{cont}}$ penalizes cross-track deviation more strongly than along-path error and tracks the path-indexed yaw and arm configurations, allowing the optimized progress $s(t)$ to absorb timing variations while preserving the geometric intent of the VLA-predicted behavior. $J_{\mathrm{prog}}$ regularizes phase distortion and progress-rate variation and discourages stagnation. $J_{\mathrm{feas}}$ penalizes violations of platform velocity, acceleration, body-rate, yaw-rate, and joint-rate limits, together with whole-body collisions. A densely sampled feasibility check is applied before commitment.

\section{Scene-Graph-Guided Scene-Scale Behavior Composition}
\label{sec:scene_graph_composition}
\subsection{Scene-Graph-Conditioned Mission Composition}
Building on the polyhedral spatial connectivity and semantic-region hierarchy of USS-Nav~\cite{gai2026ussnav}, our Scene Graph organizes object instances within an Area--Polyhedron hierarchy and encodes their spatial relations and free-space connectivity. Given a natural-language mission $\ell$, an externally hosted LLM uses this graph context and the available local VLA skills to map linguistic references to graph-supported object categories and relations, resolve implicit task dependencies, and compose an ordered graph-grounded mission plan:
\begin{equation}
\begin{aligned}
\Pi_{\ell}&=(u_1,\ldots,u_N),\\
u_i&\in\{\textsc{Navigate},\textsc{ExecuteVLA},\textsc{Recover}\}.
\end{aligned}
\label{eq:scene_graph_mission_plan}
\end{equation}
Each step refers to a graph-described task target rather than specifying metric coordinates or control commands. The sequence defines mission logic and is instantiated by a deterministic executor as a state machine, whose transitions are advanced according to navigation and VLA outcomes.

\subsection{Instance Grounding and Interaction-Region Construction}
Each target-dependent action $u_i$ carries a graph-supported relational description $\mathcal{Q}_i=(c_i,\boldsymbol{\Phi}_i)$. A deterministic resolver selects object nodes matching the category and all specified relations, and execution proceeds only when the query identifies a unique instance, i.e., $\lvert\mathcal{C}(\mathcal{Q}_i)\rvert=1$. For the resolved instance $v_i^\star$, a skill-conditioned geometric template defines an interaction region $\mathcal{I}_{\kappa_i}(v_i^\star)$ rather than a single handoff pose. Intersecting this region with clearance-eroded reachable Polyhedra and enforcing whole-body feasibility yields the candidate handoff set $\mathcal{H}_i^\star$ for subsequent transfer and VLA execution.

\subsection{Topology-Guided Whole-Body Transfer and VLA Handoff}
Each candidate handoff state $\mathbf h\in\mathcal H_i^\star$ is mapped to its containing Polyhedron. From the current cell, graph search jointly selects a terminal handoff state $\mathbf h_i^\star$ and a connected Polyhedron route $\Gamma_i=(P_0,\ldots,P_L)$ according to route cost and terminal feasibility. The selected Polyhedron sequence provides the global route structure, while its projected cell regions and inter-cell connections define an aerial corridor that constrains whole-body transfer between interaction sites. Upon arrival, control is handed to the local VLA only when the measured pose and motion satisfy the interaction-region tolerances; otherwise, the handoff is rejected.

\begin{algorithm}[t]
\caption{Scene-Graph-Grounded Mission Execution}
\label{alg:scenegraph_execution}
\small
\begin{algorithmic}[1]
\Require Mission $\ell$, Scene Graph $\mathcal G$, VLA skills $\mathcal K$
\Ensure Mission outcome $y$
\State $\Pi_\ell \gets \Call{LLMCompose}{\ell,\mathcal G,\mathcal K}$
\State $\mathcal S \gets \Call{InstantiateFSM}{\Pi_\ell}$
\While{$\mathcal S$ is not terminal}
    \State $(\mathcal Q_i,\kappa_i) \gets \Call{CurrentStep}{\mathcal S}$
    \State $(v_i^\star,\mathcal H_i^\star)
        \gets \Call{GroundAndFilter}{\mathcal Q_i,\kappa_i,\mathcal G}$
    \State $(\mathbf h_i^\star,\Gamma_i)
        \gets \Call{TopologySearch}{\mathbf x,\mathcal H_i^\star,\mathcal G}$
    \If{$v_i^\star=\varnothing$ or $\mathcal H_i^\star=\varnothing$
        or $\Gamma_i=\varnothing$}
        \State $\mathcal S \gets \Call{UpdateFSM}{\mathcal S,\textsc{Reject}}$
        \State \textbf{continue}
    \EndIf
    \State $\mathcal F_i \gets \Call{AssembleCorridor}{\Gamma_i,\mathcal G}$
    \State $r_i^{\mathrm{nav}}\gets
        \Call{ExecuteTransfer}{\mathcal F_i,\mathbf h_i^\star}$
    \If{$r_i^{\mathrm{nav}}$ and
        $\Call{VerifyHandoff}{\mathbf x,\mathbf h_i^\star}$}
        \State $r_i\gets\Call{ExecuteVLA}{\kappa_i,v_i^\star,\mathrm{MPAR}}$
    \Else
        \State $r_i\gets\textsc{TransferFailure}$
    \EndIf
    \State $\mathcal S\gets\Call{UpdateFSM}{\mathcal S,r_i}$
\EndWhile
\State \Return $y\gets\Call{Outcome}{\mathcal S}$
\end{algorithmic}
\end{algorithm}

\section{Experiments}
\label{sec:experiments}
\setlength{\textfloatsep}{8pt plus 2pt minus 1pt}
\setlength{\floatsep}{6pt plus 2pt minus 1pt}
\setlength{\intextsep}{8pt plus 2pt minus 1pt}

\subsection{Experimental Setup}

We evaluate the proposed system in MuJoCo simulation and on a physical articulated UAM. The physical platform comprises a quadrotor, a 3-DoF arm, a gripper, an NVIDIA Jetson Orin NX, an Intel RealSense D450 depth camera, and a monocular RGB camera, while VLA inference runs remotely on an NVIDIA RTX 5090. Paired evaluations use the same model, initial conditions, task configurations, and random seeds. Binary outcomes are reported as counts and rates; Table~I additionally reports 95\% Wilson confidence intervals. Continuous quantities are summarized by their median and 95th percentile (P95). Collision denotes unintended UAM contact, excluding task-prescribed gripper--object contact.

\subsection{Cross-Site Mission Capability}
\label{sec:exp_cross_site}

\begin{table}[!htbp]
  \caption{Cross-site mission performance in simulation. 
  Values denote successful trials over total trials.}
  \label{tab:cross_site}
  \centering
  \small
  \setlength{\tabcolsep}{3.5pt}
  \renewcommand{\arraystretch}{1.08}

  \begin{tabularx}{\columnwidth}{
    @{}
    >{\raggedright\arraybackslash}X
    >{\centering\arraybackslash}p{0.29\columnwidth}
    >{\centering\arraybackslash}p{0.21\columnwidth}
    @{}
  }
    \toprule
    Task & Whole-Task VLA & Ours \\
    \midrule
    Water plant             & 0/10 & 5/10 \\
    Water and discard       & 0/10 & 3/10 \\
    Discard empty bottle    & 0/10 & 6/10 \\
    Store rectangular block & 0/10 & 4/10 \\
    Store fan-shaped block  & 0/10 & 3/10 \\
    \midrule
    \textbf{Overall}
      & \textbf{0/50}
      & \textbf{21/50} \\
    Success rate
      & 0.0\%
      & 42.0\% \\
    95\% Wilson CI
      & [0.0, 7.1]\%
      & [29.4, 55.8]\% \\
    \bottomrule
  \end{tabularx}
\end{table}

We evaluate cross-site mission execution in simulation by comparing our system with a Monolithic Whole-Task VLA under matched tasks, initial conditions, and random seeds. The baseline receives the complete mission instruction and generates rolling whole-body action chunks without relational grounding, topology-guided transfer, or feasible handoff. For a controlled comparison, its outputs are realized using the same MPAR module as ours. It is trained on 1,500 complete-task trajectories comprising 231,989 action windows, 5.92$\times$ the 39,200 MuJoCo local-skill windows used by our system. The baseline uses the same initialization, LoRA recipe, input–output representation, optimization budget, and fixed-checkpoint protocol (Sec. IV), with normalization computed from its own dataset. This comparison evaluates monolithic versus compositional scene-scale execution rather than isolating policy architecture or data efficiency.

\textbf{Mission success criteria.} A cross-site episode succeeds only when an independent evaluator observes every task-specific physical milestone in the prescribed order: object acquisition followed by retained watering; acquisition, retained watering, and release into the trash bin; acquisition and release into the trash bin; or block acquisition and release into the storage box. Acquisition, watering, and storage use the same physical predicates defined in Sec.~\ref{sec:exp_progress}; trash-bin placement analogously requires explicit release followed by stable residence in its prescribed target region. A collision, payload loss, missing or misplaced release, timeout, or runtime/planning termination constitutes episode failure, and component-reported completion alone is insufficient.

As shown in Table~\ref{tab:cross_site}, the monolithic baseline completes none of the 50 trials, whereas ours completes 21/50. Its failures comprise 19 collisions and 31 runtime-planning or action-feasibility terminations. Among our 29 failed trials, one results from planning failure and 28 occur during local VLA interaction. Under oracle target and feasible-handoff conditions, the local skills succeed in 39/60 interactions, comprising 14/30 object-acquisition and 25/30 post-acquisition trials, which further identifies object acquisition as the dominant local-behavior bottleneck.

\begin{figure*}[t]
    \centering
    \includegraphics[width=\textwidth]{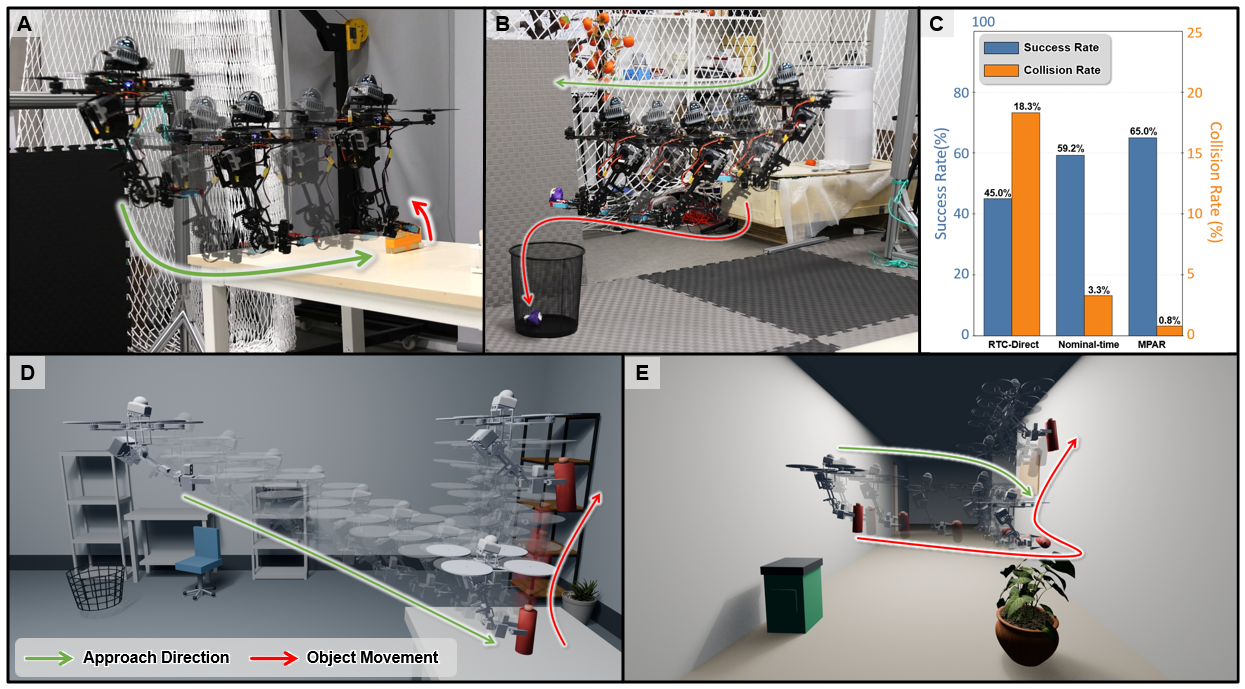}
    \caption{\textbf{Evaluation of whole-body realization.} (A--B) Representative physical local interactions under MPAR. (C) Aggregate success and collision rates for RTC-Direct, the nominal-time ablation, and MPAR. (D--E) Representative simulated storage and watering executions. Green arrows indicate the UAM approach direction, red arrows indicate object motion, and ghosted configurations show the executed whole-body trajectory.}
    \label{fig:quantitative_results}
    \vspace{-0.3cm}
\end{figure*}

\subsection{Measured-Progress Whole-Body Realization}
\label{sec:exp_progress}

\begin{table}[!htbp]
  \centering
  \caption{Task- and condition-wise success under runtime disturbances. Both views summarize the same 120 trials per method (40 per task and 30 per condition); best results are bold. Nominal-time and MPAR denote our ablation and full method, respectively.}
  \label{tab:progress_realization}

  \small
  \setlength{\tabcolsep}{2.4pt}
  \renewcommand{\arraystretch}{1.08}

  \begin{tabularx}{\columnwidth}{
    @{}
    >{\raggedright\arraybackslash}X
    >{\centering\arraybackslash}p{0.25\columnwidth}
    >{\centering\arraybackslash}p{0.25\columnwidth}
    >{\centering\arraybackslash}p{0.16\columnwidth}
    @{}
  }
    \toprule
    Task/condition & RTC-Direct~\cite{tucker2026makeitfly} & Nominal-time & \textbf{MPAR} \\
    \midrule
    \multicolumn{4}{@{}l}{\emph{Task-wise success}} \\
    Grasp Rect & 11/40 & \textbf{14/40} & 12/40 \\
    Watering   & 24/40 & 27/40 & \textbf{32/40} \\
    Storage    & 19/40 & 30/40 & \textbf{34/40} \\
    \addlinespace[1pt]
    \multicolumn{4}{@{}l}{\emph{Condition-wise success}} \\
    C0 & 18/30 & \textbf{21/30} & 19/30 \\
    C1 & 10/30 & 17/30 & \textbf{20/30} \\
    C2 & 14/30 & \textbf{17/30} & 16/30 \\
    C3 & 12/30 & 16/30 & \textbf{23/30} \\
    \bottomrule
  \end{tabularx}
\end{table}

We compare our two constrained-realization variants against RTC-Direct, a direct-execution baseline based on the RTC mechanism used by AirVLA~\cite{tucker2026makeitfly,black2025rtc}. RTC-Direct forwards RTC-conditioned action references without constrained trajectory realization. Our nominal-time ablation constructs continuous, constraint-aware MINCO trajectories but selects takeover locations according to nominal elapsed time. MPAR additionally aligns each returned chunk with measured physical progress. Because the two realization variants share the same trajectory formulation and constraints, their comparison isolates the contribution of measured-progress alignment. Unsafe publication denotes committing a MINCO trajectory that fails the feasibility check in Sec.~V-C; it is inapplicable to RTC-Direct.

We evaluate all methods on rectangular-block grasping, plant watering, and storage placement under four controlled asynchronous-execution conditions, denoted C0--C3. C0 adds no artificial delay beyond native inference and communication; C1 and C2 delay each completed action chunk by an additional 250 and 500~ms, respectively. C3 augments C2 with one 250-ms hold of the latest base and arm commands per episode, emulating a transient command-update stall while physics and state estimation continue normally. These conditions progressively decouple elapsed time from physical execution progress. We conduct ten paired trials per task, method, and condition, yielding 360 trials.

\textbf{Success criteria.} Task success is determined from measured simulator states rather than workflow completion alone. Grasp Rect succeeds when, after gripper closure, the correct block is lifted clear of the tabletop while simultaneously contacting both finger pads with a normal force of at least $0.20$~N on each side. Watering succeeds when the bottle remains grasped and the measured end effector stays within $0.22$~m of the flower-pot center while satisfying $\theta_3\geq\pi/2-0.03$~rad continuously for at least $3.0$~s. Together with the calibrated tool--bottle geometry, this spatial condition keeps the bottle mouth above the pot during pouring. Storage succeeds when the correct block is explicitly released and, at least $0.35$~s after opening the gripper, its center satisfies $|x-x_{\mathrm{box}}|\leq0.20$~m and $|y-y_{\mathrm{box}}|\leq0.22$~m, with object speed no greater than $0.15$~m/s.

As shown in Table~\ref{tab:progress_realization}, MPAR succeeds in 78/120 trials (65.0\%), compared with 71/120 (59.2\%) for our nominal-time ablation and 54/120 (45.0\%) for RTC-Direct. Under the predesignated C2--C3 conditions, the corresponding rates are 65.0\%, 55.0\%, and 43.3\%. The largest separation occurs under C3, where MPAR reaches 23/30 successes, versus 16/30 and 12/30, respectively. The task-level effect is not universal: MPAR improves Watering and Storage but does not outperform the nominal-time ablation on Grasp Rect.

\begin{table}[!htbp]
  \caption{Mechanism-level descriptive comparison of nominal-time alignment and MPAR. Entries report task-balanced C2/C3 aggregates defined in the text; rejection is pooled across C2--C3. Lower is better.}
  \label{tab:mpar_mechanism}
  \centering
  \small
  \setlength{\tabcolsep}{3.5pt}
  \renewcommand{\arraystretch}{1.08}

  \begin{tabularx}{\columnwidth}{
    @{}
    >{\raggedright\arraybackslash}X
    >{\centering\arraybackslash}p{0.23\columnwidth}
    >{\centering\arraybackslash}p{0.20\columnwidth}
    >{\centering\arraybackslash}p{0.21\columnwidth}
    @{}
  }
    \toprule
    Metric & Nominal-time & MPAR & Reduction \\
    \midrule
    Phase P50 [s]
      & 0.510/0.459 
      & \textbf{0.310/0.273} 
      & 39.3/40.5\% \\

    Phase P95 [s]
      & 0.830/0.831 
      & \textbf{0.444/0.508} 
      & 46.6/38.9\% \\

    Skipped arc [cm]  
      & 8.8/8.7     
      & \textbf{3.7/3.1}     
      & 58.0/64.4\% \\

    Repeated arc [cm] 
      & 12.1/10.3   
      & \textbf{7.8/5.2}     
      & 35.5/49.5\% \\

    Rejection [\%]    
      & 17.42        
      & \textbf{12.81}       
      & 26.5\% \\
    \bottomrule
  \end{tabularx}
\end{table}

Metrics are aggregated episode-first and task-balanced: phase-error P50/P95 are computed over valid handoffs within each episode, while skipped and repeated arc lengths are summed; each metric is then summarized by the median over ten episodes per task and the median across the three tasks.

The mechanism measurements in Table~\ref{tab:mpar_mechanism} further characterize the effect of progress alignment. Across 60 paired C2--C3 episodes, MPAR reduces episode-level P50 and P95 phase errors by an average of 0.212 and 0.315~s; their paired 95\% confidence intervals, $[0.163,0.262]$ and $[0.234,0.393]$~s, exclude zero. It also reduces episode-level skipped and repeated path lengths by 7.44 and 1.89~cm, respectively, while decreasing candidate-trajectory rejection from 17.42\% to 12.81\%. These gains are not accompanied by additional safety violations: collision-bearing runs decrease from 22/120 for RTC-Direct and 4/120 for our nominal-time ablation to 1/120 for MPAR, with no unsafe publications for either constrained variant. Representative physical and simulated executions are shown in Fig.~\ref{fig:quantitative_results}.

\subsection{Causal Ablation of the Scene-Graph Physical Interface}
\label{sec:exp_scenegraph}

\begin{table}[!htbp]
  \centering
  \caption{Scene Graph physical-interface ablation. $-$G, $-$R, and $-$H remove relational grounding, topology-guided routing, and feasible handoff, respectively. Store, Water, and W$\rightarrow$D (Water then Discard) are task-wise mission successes; Mission is their aggregate.}
  \label{tab:scenegraph_ablation}

  \small
  \setlength{\tabcolsep}{2.5pt}
  \renewcommand{\arraystretch}{1.04}

  \begin{tabularx}{\columnwidth}{
    @{}
    >{\raggedright\arraybackslash}X
    *{4}{>{\centering\arraybackslash}p{0.155\columnwidth}}
    @{}
  }
    \toprule
    Metric & $-$G & $-$R & $-$H & Full \\
    \midrule

    \multicolumn{5}{@{}l}{\emph{Stage-level success}} \\
    Grounding & 20/35 & \textbf{35/35} & \textbf{35/35} & \textbf{35/35} \\
    Transfer  & N/A   & 27/34          & \textbf{35/35} & 34/35 \\
    Handoff   & N/A   & 26/27          & 26/35          & \textbf{34/34} \\
    Local     & N/A   & 16/26          & 15/26          & \textbf{26/34} \\

    \addlinespace[1pt]
    \multicolumn{5}{@{}l}{\emph{Episode-level outcomes}} \\
    Interface      & 0/15 & 8/15 & 7/15 & \textbf{14/15} \\
    Store          & 0/5  & 1/5  & 0/5  & \textbf{2/5} \\
    Water          & 0/5  & 2/5  & 1/5  & \textbf{4/5} \\
    W$\rightarrow$D & 0/5  & 1/5  & 1/5  & \textbf{3/5} \\
    Mission        & 0/15 & 4/15 & 2/15 & \textbf{9/15} \\
    Nav. collision & N/A  & 4/15 & \textbf{1/15} & \textbf{1/15} \\

    \bottomrule
  \end{tabularx}

  \vspace{0.5mm}
  \parbox{\columnwidth}{%
    \footnotesize\emph{Notes:} Local is event-level and may include multiple failures per episode; N/A denotes a stage not entered.
  }
\end{table}

We evaluate the Scene Graph physical interface across three cross-site mission types, with five paired episodes per task and variant. We compare the system with three single-component ablations: removing relational grounding, removing topology-guided routing, and replacing feasible-region handoff with a nominal point target. All variants use identical mission instructions, scene configurations, local VLA policies, MPAR realization, and fixed LLM-generated task sequences. Grounding is evaluated over all target references; other stage metrics are conditioned on entry. Each local event is one VLA skill invocation attempted once; after failure, later events proceed only if execution remains safe, without external reinitialization. For $-G$, all queries are audited before motion, and downstream stages are entered only if every query resolves the correct unique instance. Interface success, mission success, and navigation collision are episode-level; mission success additionally requires every prescribed local event.

As shown in Table~\ref{tab:scenegraph_ablation}, the complete system achieves interface success in 14/15 episodes and mission success in 9/15. Without relational grounding, no episode resolves all targets; removing routing increases navigation collisions, while nominal-point handoffs reduce handoff success. The task-wise breakdown covers sources of difficulty: Store Rect is dominated by object acquisition, whereas Water then Discard requires a longer three-interaction sequence.

\subsection{Physical Full-Stack Validation}
\label{sec:exp_physical}

Under oracle target and feasible-handoff conditions, the local VLA completes bottle grasping, rectangular-block grasping, watering, and storage placement in 2/5, 1/5, 3/5, and 3/5 trials, respectively. These results demonstrate physical executability without demonstrations collected on the target UAM.

The full system also succeeds in 2/5 physical trials of the cross-site Water Plant mission.

\section{Conclusion and Future Work}
This work unifies synthetic whole-body VLA training, measured-progress-aligned realization, and scene-graph-guided composition for scene-scale aerial manipulation, with simulation and physical validation. The current system relies on a preconstructed Scene Graph registered to a metric map and remote VLA inference, while its mission-level performance remains limited by individual local skills. Future work will investigate online Scene Graph construction and update, onboard inference, closed-loop failure processing, and broader physical-interaction capabilities.

\bibliography{ICRA2022}

@string{icra = {Proc. of the {IEEE} Intl. Conf. on Robot. and Autom.}}

@string{iros = {Proc. of the {IEEE/RSJ} Intl. Conf. on Intell. Robots and Syst.}}

@inproceedings{zitkovich2023rt2,
  title={{RT-2}: Vision-Language-Action Models Transfer Web Knowledge to Robotic Control},
  author={Zitkovich, Brianna and Yu, Tianhe and Xu, Sichun and Xu, Peng and Xiao, Ted and Xia, Fei and Wu, Jialin and Wohlhart, Paul and Welker, Stefan and Wahid, Ayzaan and others},
  booktitle={Proceedings of the Conference on Robot Learning},
  volume={229},
  pages={2165--2183},
  year={2023}
}

@inproceedings{kim2024openvla,
  title={{OpenVLA}: An Open-Source Vision-Language-Action Model},
  author={Kim, Moo Jin and Pertsch, Karl and Karamcheti, Siddharth and Xiao, Ted and Balakrishna, Ashwin and Nair, Suraj and Rafailov, Rafael and Foster, Ethan P. and Sanketi, Pannag R. and Vuong, Quan and others},
  booktitle={Proceedings of the Conference on Robot Learning},
  volume={270},
  pages={2679--2713},
  year={2025}
}

@inproceedings{openx2024,
  title={Open {X-E}mbodiment: Robotic Learning Datasets and {RT-X} Models},
  author={{Open X-Embodiment Collaboration}},
  booktitle={IEEE International Conference on Robotics and Automation},
  year={2024}
}

@inproceedings{sun2026airvla,
  title={{AIR-VLA}: Vision-Language-Action Systems for Aerial Manipulation},
  author={Sun, Jianli and Tian, Bin and Zhang, Qiyao and Li, Chengxiang and Song, Zihan and Cui, Zhiyong and Lv, Yisheng and Tian, Yonglin},
  booktitle={Proceedings of the 43rd International Conference on Machine Learning},
  series={Proceedings of Machine Learning Research},
  volume={306},
  pages={117222--117240},
  year={2026},
  publisher={PMLR}
}

@article{tucker2026makeitfly,
  title={{$\pi$}, But Make It Fly: Physics-Guided Transfer of {VLA} Models to Aerial Manipulation},
  author={Tucker, Johnathan and Liu, Denis and Swann, Aiden and Ren, Allen and Yu, Javier and Sun, Jiankai and Kim, Brandon and McGranahan, Lachlain and Vuong, Quan and Schwager, Mac},
  journal={arXiv preprint arXiv:2603.25038},
  year={2026}
}

@article{sun2026autofly,
  title={{AutoFly}: Vision-Language-Action Model for {UAV} Autonomous Navigation in the Wild},
  author={Sun, Xiaolou and Si, Wufei and Ni, Wenhui and Li, Yuntian and Wu, Dongming and Xie, Fei and Guan, Runwei and Xu, He-Yang and Ding, Henghui and Wu, Yuan and Yue, Yutao and Huang, Yongming and Xiong, Hui},
  journal={arXiv preprint arXiv:2602.09657},
  year={2026}
}

@article{gupta2025umi,
  title={{UMI-on-Air}: Embodiment-Aware Guidance for Embodiment-Agnostic Visuomotor Policies},
  author={Gupta, Harsh and Guo, Xiaofeng and Ha, Huy and Pan, Chuer and Cao, Muqing and Lee, Dongjae and Scherer, Sebastian and Song, Shuran and Shi, Guanya},
  journal={arXiv preprint arXiv:2510.02614},
  year={2025}
}

@inproceedings{black2025rtc,
  title={Real-Time Execution of Action Chunking Flow Policies},
  author={Black, Kevin and Galliker, Manuel Y. and Levine, Sergey},
  booktitle={Advances in Neural Information Processing Systems},
  year={2025}
}

@inproceedings{he2025flying,
  title={{Flying Hand}: End-Effector-Centric Framework for Versatile Aerial Manipulation Teleoperation and Policy Learning},
  author={He, Guanqi and Guo, Xiaofeng and Tang, Luyi and Zhang, Yuanhang and Mousaei, Mohammadreza and Xu, Jiahe and Geng, Junyi and Scherer, Sebastian and Shi, Guanya},
  booktitle={Proceedings of Robotics: Science and Systems},
  year={2025}
}

@inproceedings{wang2026remac,
  title={Real-Time Robot Execution with Masked Action Chunking},
  author={Wang, Haoxuan and Zhang, Gengyu and Yan, Yan and Shang, Yuzhang and Kompella, Ramana Rao and Liu, Gaowen},
  booktitle={International Conference on Learning Representations},
  year={2026}
}

@article{jiang2026futurertc,
  title={{FutureRTC}: Real-Time Robot Execution with Anticipatory-Conditioned Action Chunking},
  author={Jiang, Hai and Zou, Yixian and Liang, Binbin and Liu, Boqian and Meng, Fanman and Liu, Shuaicheng},
  journal={arXiv preprint arXiv:2607.24008},
  year={2026}
}

@article{zhepei22tro,
  title={Geometrically Constrained Trajectory Optimization for Multicopters},
  author={Wang, Zhepei and Zhou, Xin and Xu, Chao and Gao, Fei},
  journal={IEEE Transactions on Robotics},
  volume={38},
  number={5},
  pages={3259--3278},
  year={2022}
}

@article{deng25tro,
  title={Whole-Body Integrated Motion Planning for Aerial Manipulators},
  author={Deng, Weiliang and Chen, Hongming and Ye, Biyu and Chen, Haoran and Li, Ziliang and Lyu, Ximin},
  journal={IEEE Transactions on Robotics},
  volume={41},
  pages={6661--6679},
  year={2025}
}

@inproceedings{gu2024conceptgraphs,
  title={{ConceptGraphs}: Open-Vocabulary {3D} Scene Graphs for Perception and Planning},
  author={Gu, Qiao and Kuwajerwala, Alihusein and Morin, Sacha and Jatavallabhula, Krishna Murthy and Sen, Bipasha and Agarwal, Aditya and Rivera, Corban and Paul, William and Ellis, Kirsty and Chellappa, Rama and Gan, Chuang and de Melo, Celso Miguel and Tenenbaum, Joshua B. and Torralba, Antonio and Shkurti, Florian and Paull, Liam},
  booktitle={2024 IEEE International Conference on Robotics and Automation (ICRA)},
  pages={5021--5028},
  year={2024},
  organization={IEEE}
}

@inproceedings{werby2024hovsg,
  title={Hierarchical Open-Vocabulary {3D} Scene Graphs for Language-Grounded Robot Navigation},
  author={Werby, Abdelrhman and Huang, Chenguang and B{\"u}chner, Martin and Valada, Abhinav and Burgard, Wolfram},
  booktitle={Robotics: Science and Systems},
  year={2024}
}

@inproceedings{rana2023sayplan,
  title={{SayPlan}: Grounding Large Language Models Using {3D} Scene Graphs for Scalable Robot Task Planning},
  author={Rana, Krishan and Haviland, Jesse and Garg, Sourav and Abou-Chakra, Jad and Reid, Ian and S{\"u}nderhauf, Niko},
  booktitle={Conference on Robot Learning},
  year={2023}
}

@article{yan2025dovsg,
  title={Dynamic Open-Vocabulary {3D} Scene Graphs for Long-Term Language-Guided Mobile Manipulation},
  author={Yan, Zhijie and Li, Shufei and Wang, Zuoxu and Wu, Lixiu and Wang, Han and Zhu, Jun and Chen, Lijiang and Liu, Jihong},
  journal={IEEE Robotics and Automation Letters},
  year={2025}
}

@article{gai2026ussnav,
  title={{USS-Nav}: Unified Spatio-Semantic Scene Graph for Lightweight {UAV} Zero-Shot Object Navigation},
  author={Gai, Weiqi and Gao, Yuman and Zhou, Yuan and Xie, Yufan and Liu, Zhiyang and Wu, Yuze and Zhou, Xin and Gao, Fei and Meng, Zhijun},
  journal={arXiv preprint arXiv:2602.00708},
  year={2026}
}

@inproceedings{lin2025float,
  title={{FLOAT} Drone: A Fully-actuated Coaxial Aerial Robot for Close-Proximity Operations},
  author={Lin, Junxiao and Ji, Shuhang and Wu, Yuze and Wu, Tianyue and Han, Zhichao and Gao, Fei},
  booktitle={2025 IEEE/RSJ International Conference on Intelligent Robots and Systems (IROS)},
  pages={7216--7223},
  year={2025},
  organization={IEEE}
}

@article{ductam,
      title={{DuctAM}: A Duct-Assisted Quadrotor-Based Aerial Manipulator Enabling High-Force Push-and-Pull Interactions}, 
      author={Yi Wang and Rui Jin and Xinhang Xu and Haotian Jin and Ruiyang Liu and Yizhuo Yang and Lihua Xie},
      journal={arXiv preprint arXiv:2609.15861},
      year={2026},
}

@article{jinjie2026tro,
      title={Effector-Centric {NMPC} of Tiltable-Multirotors for Offset-Free Omnidirectional Aerial Manipulation}, 
      author={Jinjie Li and Yicheng Chen and Johannes Kübel and Haokun Liu and Junichiro Sugihara and Moju Zhao},
      journal={arXiv preprint arXiv:2608.17819},
      year={2026},
}

@article{piwebsite2025pi05,
  title={{$\pi_{0.5}$}: A Vision-Language-Action Model with Open-World Generalization},
  author={{Physical Intelligence} and others},
  journal={arXiv preprint arXiv:2504.16054},
  year={2025}
}

@inproceedings{gsplanner,
  author    = {Jin, Rui and Gao, Yuman and Wang, Yingjian
               and Wu, Yuze and Lu, Haojian and Xu, Chao
               and Gao, Fei},
  title     = {{GS-Planner}: A Gaussian-Splatting-based Planning
               Framework for Active High-Fidelity Reconstruction},
  booktitle = {Proc. IEEE/RSJ International Conference on
               Intelligent Robots and Systems (IROS)},
  pages     = {11202--11209},
  year      = {2024}
}

@article{sketchprior,
  author = {Xu, Xinhang and Liu, Ruiyang and Jin, Haotian
            and Wang, Yi and Shen, Hongming
            and Li, Jianping and Xie, Lihua},
  title  = {From Sketch Prior to Trajectories: A Mission-Oriented Coordinated Navigation Framework for Indoor {UAV} Swarm},
      journal={arXiv preprint arXiv:2607.11386},
      year={2026},
}

@article{quadhand,
      title={{QuadHand}: A Compact Quadrotor Aerial Manipulator with {MRC-SDF}-Based Whole-Body Motion Planning}, 
      author={Rui Jin and Ruiyang Liu and Xinhang Xu and Haotian Jin and Yi Wang and Yizhuo Yang and Lihua Xie},
      journal={arXiv preprint arXiv:2609.35094},
      year={2026},
}

@article{tethered,
author = {Rui Jin  and Xinhang Xu  and Yizhuo Yang  and Jianping Li  and Muqing Cao  and Lihua Xie },
title = {Tethered {UAV} Autonomous Knotting on Environmental Structures for Transport},
journal = {Cyborg and Bionic Systems},
volume = {6},
number = {},
pages = {0450},
year = {2025},
doi = {10.34133/cbsystems.0450},
}
\end{document}